\documentclass[letterpaper, 10 pt, journal, twoside]{IEEEtran}
\IEEEoverridecommandlockouts
\usepackage{amsmath,amsfonts}
\usepackage{algorithmic}
\usepackage{algorithm}
\usepackage{array}
\usepackage[caption=false,font=normalsize,labelfont=sf,textfont=sf]{subfig}
\usepackage{textcomp}
\usepackage{stfloats}
\usepackage{url}
\usepackage{verbatim}
\usepackage{graphicx}
\usepackage{cite}
\usepackage{booktabs}
\usepackage{orcidlink}
\usepackage{bm}
\usepackage{multirow}
\usepackage{float}
\usepackage{amstext}
\usepackage{amsthm}
\usepackage{mathrsfs}
\usepackage{makecell}
\usepackage{colortbl}
\usepackage{hyperref}
\usepackage{amssymb}
\title{\LARGE \bf SD4R: Sparse-to-Dense Learning \\ for 3D Object Detection with 4D Radar}

\author{Xiaokai Bai\textsuperscript{*}, Jiahao Cheng\textsuperscript{*}, Songkai Wang, Yixuan Luo, Lianqing Zheng, Xiaohan Zhang\\ Si-Yuan Cao, and Hui-Liang Shen\textsuperscript{\dag}, \emph{Senior Member, IEEE}
\thanks{This work was supported in part by the National Key Research and Development Program of China under grant 2023YFB3209800, in part by the Aeronautical Science Foundation of China under grant 2024M071076001, and in part by the National Natural Science Foundation of China under grant 62301484.}
\thanks{\textsuperscript{*}Equal Contribution, \textsuperscript{\dag}Corresponding author.}

\thanks{Xiaokai Bai, Jiahao Cheng, Songkai Wang, Yixuan Luo and Hui-Liang Shen are with the College of Information Science and Electronic Engineering, Zhejiang University, Hangzhou 310027, China (e-mail: shawnnnkb@zju.edu.cn, 3230105428@zju.edu.cn, 3230105116@zju.edu.cn, 3240102341@zju.edu.cn, zhangxh2023@zju.edu.cn, shenhl@zju.edu.cn).}
\thanks{Si-Yuan Cao is with the Ningbo Global Innovation Center, Zhejiang University, Ningbo 315100, China (e-mail: cao\_siyuan@zju.edu.cn).}
\thanks{Lianqing Zheng is with the School of Automotive Studies, Tongji University, Shanghai 201804, China (e-mail: zhenglianqing@tongji.edu.cn).}	

}

\begin{document}
\maketitle
\markboth{The IEEE International Conference on Intelligent Transportation Systems (ITSC).}
{Bai \MakeLowercase{\textit{et al.}}: SD4R: Sparse-to-Dense Learning for 3D Object Detection with 4D Radar}

\vspace{-10pt}
\begin{abstract}
4D radar measurements offer an affordable and weather-robust solution for 3D perception. However, the inherent sparsity and noise of radar point clouds present significant challenges for accurate 3D object detection, underscoring the need for effective and robust point clouds densification. Despite recent progress, existing densification methods often fail to address the extreme sparsity of 4D radar point clouds and exhibit limited robustness when processing scenes with a small number of points. In this paper, we propose SD4R, a novel framework that transforms sparse radar point clouds into dense representations. SD4R begins by utilizing a foreground point generator (FPG) to mitigate noise propagation and produce densified point clouds. Subsequently, a logit-query encoder (LQE) enhances conventional pillarization, resulting in robust feature representations. Through these innovations, our SD4R demonstrates strong capability in both noise reduction and foreground point densification. Extensive experiments conducted on the publicly available View-of-Delft dataset demonstrate that SD4R achieves state-of-the-art performance. Source code is available at \href{https://github.com/shawnnnkb/LGDD}{\textcolor[RGB]{213, 43, 107}{https://github.com/lancelot0805/SD4R}}.

\end{abstract}


\section{Introduction}
3D object detection is essential for applications like autonomous driving and robotics, as it enables accurate scene understanding and target localization through 3D bounding box estimation based on sensor data \cite{PointNet, VoxelNet}. While cameras capture rich texture and color, they lack direct depth measurement capabilities \cite{LSS, BEVdepth}. LiDAR, on the other hand, provides high-resolution 3D point clouds for precise depth perception \cite{LidarDetection}, but its high cost limits its widespread use. Additionally, both cameras and LiDAR are vulnerable to adverse weather, leading to performance degradation. Recently, 4D radar has emerged as an affordable and weather-robust alternative, providing measurements of range, azimuth, elevation, and velocity \cite{TJ4DRadSet,VoD, OmniHD, Doracamom, RaGS, Wavelet}. 4D radar has rapidly advanced as a promising sensor.

\begin{figure}[ht]
    \centering
    \includegraphics[width=\linewidth]{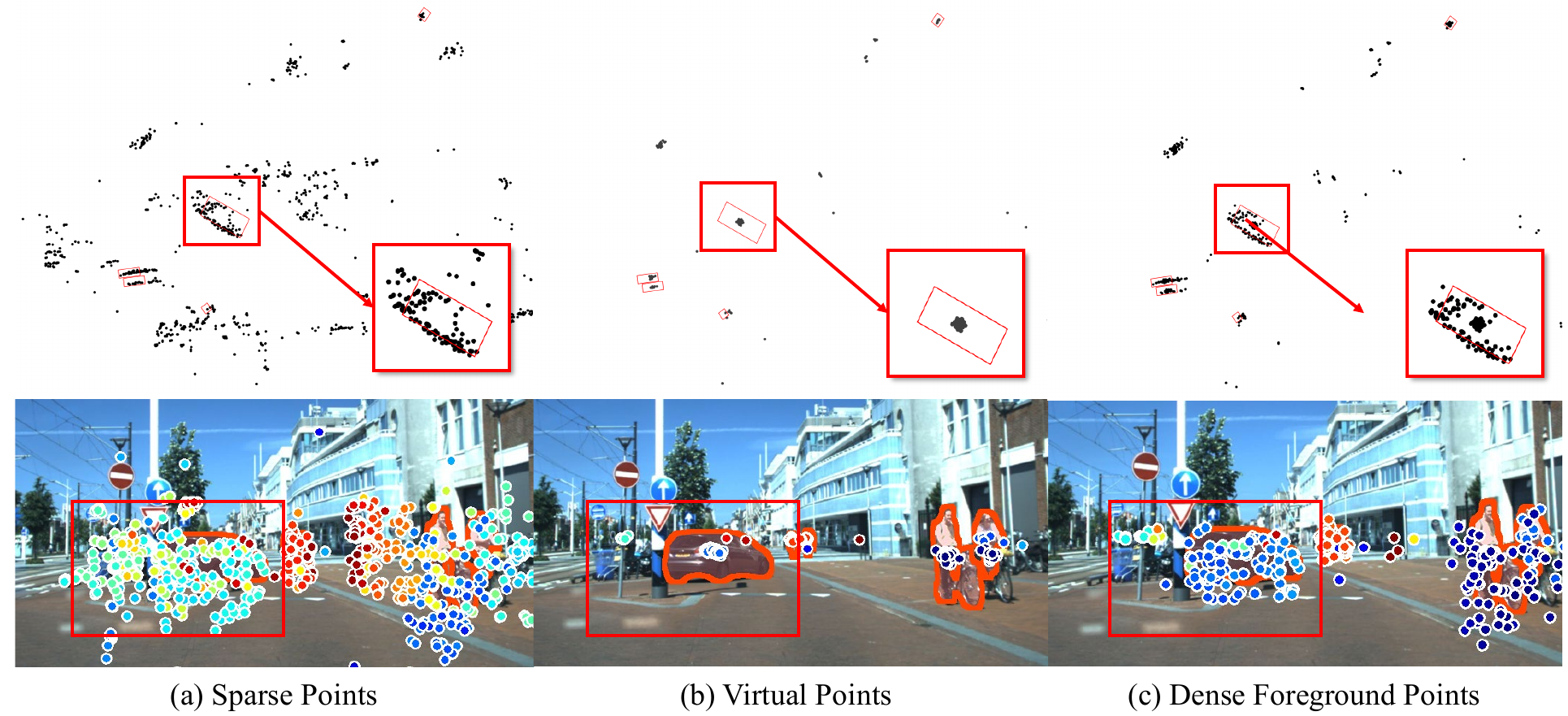}
    \caption{Performance of our SD4R in real-scene 3D detection. The sparse points refer to the original radar measurements, while the virtual points are generated only based on the foreground points. The second row shows the corresponding visualization results. Our SD4R framework demonstrates strong capability in both noise reduction and foreground point densification.  }
    \label{fig:head}
\end{figure}

However, 4D radar measurements are sparse, especially in foreground regions, posing challenges for accurate 3D object detection \cite{sparse}. To address this, transforming sparse point clouds into dense ones by generating virtual points has become a key research area. Two main challenges arise in this process. First, noisy points must be identified and excluded to prevent noise propagation; second, virtual points need to be generated from sparse point clouds to increase their density and improve detection quality (see Figure. \ref{fig:head}).

For noise reduction, Voxel-based noise reduction techniques often discard valuable information \cite{RPFA-Net, PointPillars}.  Virtual point generation methods are also suboptimal. They are divided into multi-modal and single-modal types. Multi-modal approaches rely on cameras, inheriting their sensitivity to weather \cite{multi-virtual1, multi-virtual3}, prompting the development of radar-only models. Single-modal approaches, such as \cite{PCRGNN, SIE}, use a two-stage process, first generating 3D proposals and then refining them to generate virtual points. These approaches suit dense LiDAR point clouds, where proposals contain enough points for spatial information, but struggle with 4D radar. The sparsity of radar data leads to inaccurate proposals and insufficient information, making LiDAR-based techniques suboptimal for 4D radar.

To address these challenges, we propose SD4R, a novel framework designed to transform sparse point clouds into dense ones. Specifically, we introduce a foreground point generator (FPG) that generates virtual points directly from raw point clouds, bypassing the need for proposal-dependent sampling and effectively mitigating the sparsity issue. Simultaneously, we evaluate the category likelihood of each point to identify and exclude noisy points during the virtual point generation process, thereby preventing noise propagation. Additionally, we present a logit-query encoder (LQE) that leverages category probabilities to enhance pillar representations. Finally, the detection results are obtained through a 3D detection head. The main contributions of this work are summarized as follows:
\begin{itemize}
    \item We introduce SD4R, a novel framework designed to addresses the challenges of noise and sparsity, effectively transforming sparse point clouds into dense ones.
    \item We devise the foreground points generator (FPG), which generates virtual foreground points to densify sparse point clouds and evaluates the category likelihood to mitigate the propagation of noise.
    \item We design the logit-query encoder (LQE) that utilizes category probabilities to enhance pillar features, resulting in more robust and comprehensive feature representations.
    \item  Experiments conducted on the publicly available View-of-Delft dataset demonstrate the effectiveness of SD4R, achieving state-of-the-art performance.
\end{itemize}



\section{Related Work}
\subsection{3D Object Detection with 4D Radar}
Approaches for 3D object detection using 4D radar point clouds are generally classified into point-based \cite{PointRCNN}, pillar-based \cite{PointPillars}, and voxel-based \cite{VoxelNet}. Point-based approaches like Radar PointGNN \cite{RPGNN}, which utilized GNN \cite{GNN} to effectively extract features from sparse radar point clouds through a graph representation, treating each radar point independently, allowing for flexible representation of the point cloud. However, point-based approaches can be computationally expensive, as they require complex operations for feature extraction from sparse and noisy radar data. 

The pillar-based approach \cite{LGDD}, in contrast, strikes a balance between computational efficiency and feature extraction. This approach divides the radar point cloud into pillars and processes the points within each pillar, reducing the complexity associated with point-based and voxel-based methods. It has gained popularity in 4D radar detection because it is highly computationally efficient and better handles the sparse and noisy nature of radar point clouds \cite{pillars-for-radar}. In early studies, such as \cite{VoD}, the PointPillars \cite{PointPillars} was tested on the View-of-Delft (VoD) dataset, where pillar-grid parameters were adjusted for radar sensors. RadarPillarNet \cite{RCFusion} further tailors the pillar network to radar inputs by replacing generic PFN layers with radar-specific modules, thereby extracting more discriminative attributes. RadarPillars \cite{RadarPillars} exploits radial velocity measurements and integrates a PillarAttention block to selectively emphasize informative returns, improving overall feature quality. However, they still face the limitation of the sparsity of point clouds and absence of central points.

\subsection{Point Cloud Completion }
The task of densifying sparse point clouds through the generation of virtual points is critical for downstream perception tasks. Existing methodologies can be broadly categorized into multi-modal and single-modal paradigms.  Multi-modal approaches such as \cite{multi-virtual1,multi-virtual3,SARCD}, aim to leverage the rich semantic and texture information from camera images to guide point cloud completion. This strategy generates contextually-aware virtual points, potentially even embedding semantic labels. However, their fundamental reliance on optical cameras renders them highly susceptible to performance degradation under adverse weather conditions. Single-modal approaches such as \cite{PCRGNN,SIE}, designed for LiDAR, adopt a two-stage pipeline. These approaches first generate initial bounding box proposals via a region proposal network and then process the point clouds within these proposals to produce additional virtual points. While effective for dense LiDAR point clouds, where proposals contain sufficient points to leverage rich spatial information, these approaches struggle in 4D radar scenarios. The limited number of points in initial proposals in sparse 4D radar data hinders accurate shape recovery by generation modules. Moreover, sparse point clouds lead to positioning errors in proposals. Moreover, some approaches \cite{multiframesinput, multiframeinput2, beyond} reconstruct point clouds through multi-frame input but suffer from high computation cost. In this work, we introduce a direct center-point voting mechanism operating on a single radar frame. Each point votes to predict a common object center, a process that is designed to be effective even with the highly sparse point clouds characteristic of 4D radar. 

\begin{figure*}
\centering
\includegraphics[width=0.9\linewidth]{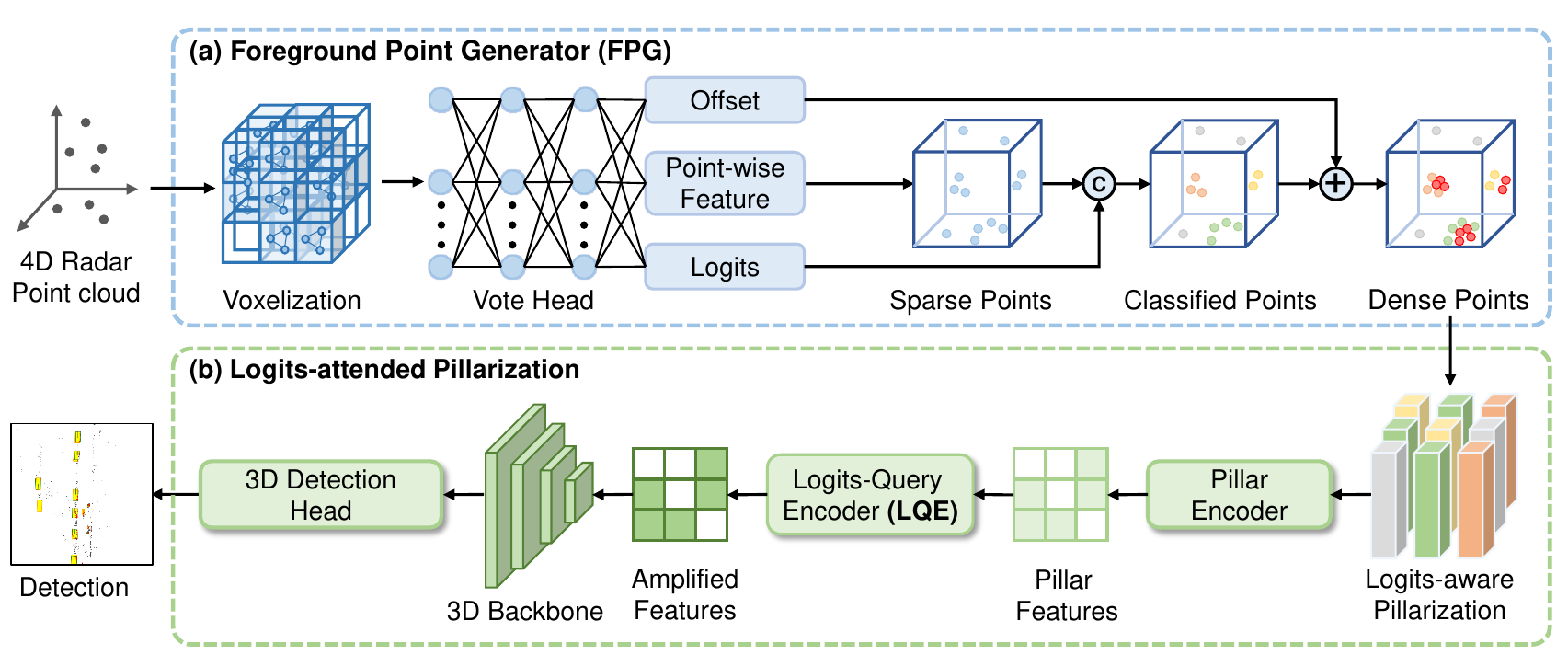}
\caption{(a) The proposed SD4R pipeline begins with voxelization of 4D radar point cloud, followed by processing through the VoteHead. This step predicts offsets between points and their corresponding object centers, classification logits, and point-wise features. These point-wise features are then concatenated with the logits to classify the point clouds. Subsequently, virtual points are generated at positions determined by the offsets, resulting in a densified point cloud. (b) The densified point cloud undergoes pillarization \cite{RCFusion} to extract features. To further address the sparsity of radar data, we introduce a logit-query encoder (LQE) module, which aggregates features from neighboring points into pillars, leading to more robust representations. Finally, the detection head processes these features to generate the final detection outputs. }
\label{fig:architecture}
\end{figure*}

\section{Method}
\subsection{Overview}
We present the overall pipeline of our SD4R designed for transformation from sparse point clouds to dense point clouds in Fig~\ref{fig:architecture}. The proposed method is structured into two main stages. Initially, virtual points are generated from existing points to produce denser foreground points in the foreground point generator as detailed in Section \ref{FPG}. Subsequently, features are extracted from the densified point clouds, and detection results are derived through the logits-attended pillarization as detailed in Section \ref{feature}.

\subsection{Foreground Point Generator}\label{FPG}
To densify the sparse point clouds, we implement a voting mechanism to estimate object centers, where we generate virtual points. Different from Lidar-based approaches, our direct voting mechanism is designed to be more robust to the extreme sparsity of 4D radar data.  Firstly, to mitigrate the impact of noise, we employ a traditional voxelization network to encode the original point cloud into voxels. Subsequently, voxel-level features are mapped back to point-level features by integrating voxel representations with the spatial offsets between points and their corresponding voxel centroids, effectively suppresses noise while preserving essential point information, resulting in robust point-wise feature.

From each point-wise feature, we employ a multi-layer perceptron (MLP) within the vote head to predict semantic logits and offsets. Each point yields a logit vectoir of dimension ${K}$, corresponding to the number of classes (e.g., pedestrian, cyclist, car, noise), and an offset vector of dimension ${3K}$. For every point, the offset is computed for $K$ classes. 

To obtain classified points and filter out noisy points, the logits of the $i$-th point are converted to class probabilities using the softmax function.
\begin{equation}
P_{i}^c = \frac{\exp(l_{i}^c)}{\sum_{c'=1}^{K}\exp(l_{i}^{c'})}, 
\quad c=1,\dots,K,
\end{equation}
where $l_{i}^c$ denotes the logit for class $c$, and $P_{i}^c$ s the predicted probability for that class. The foreground confidence is defined as 
\begin{equation}
\pi_{i} = 1 - P_{i}^{k},
\end{equation}
where $P_{i}^{k}$ represents the probability of the background class. Points with $\pi_{i}>\tau$ are retained as foreground points, effectively filtering out noise. For each foreground point, we select the most likely class $c_i$ based on the highest $P_{i}^c$ , yielding classified points 

\begin{figure}[!t]
  \centering
  \includegraphics[
    width=\columnwidth,
    height=0.4\textheight,
    keepaspectratio
  ]{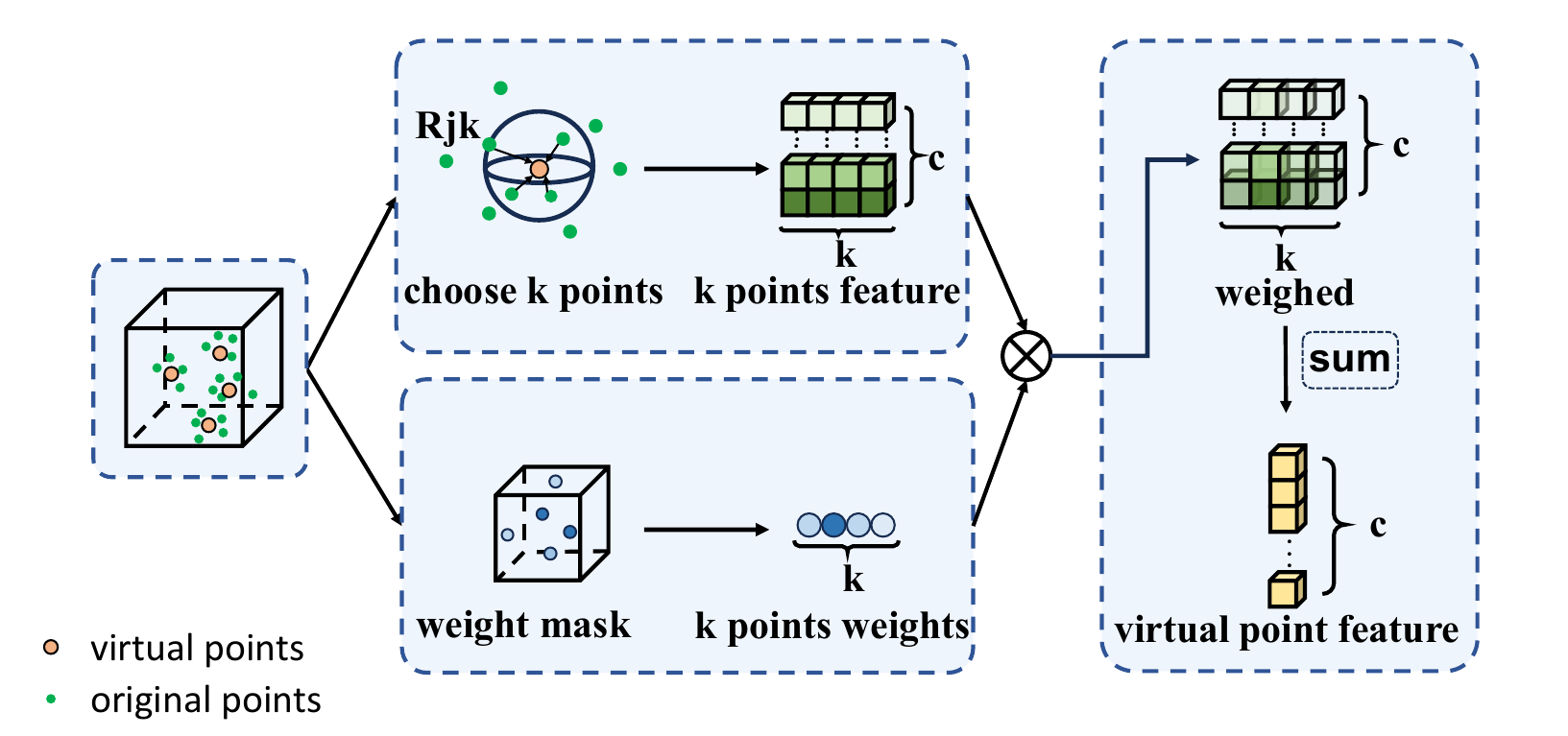}
  \caption{Based on each virtual point, we select \(k\) original points. The upper branch conducts the selection and feature extraction of \(k\) points, while the lower branch assigns weights to \(k\) points based on distance. Finally, the features of \(k\) points are multiplied by their weights and summed to obtain the feature of the virtual point.}
  \label{fig:KNN}
\end{figure}

To generate dense points, we extract the corresponding 3D offset  $\mathbf{o}_i \in \mathbb{R}^3$ for each foreground point and compute the coordinates of the $i$-th virtual point using the original point’s coordinates   $\mathbf{p}_i$
\begin{equation}
\mathbf{v}_i = \mathbf{p}_i + \mathbf{o}_i
,
\mathbf{v}_i \in \mathbb{R}^3.
\end{equation}
Next, we identify the $k$ nearest original points to each virtual point based on Euclidean distance, expressed as \begin{equation}
D_{ij} = \bigl\lVert \mathbf{v}_i - \mathbf{p}_j \bigr\rVert_2,
\quad
\forall\,\mathbf{p}_j \in \mathcal{P}_{\mathrm{raw}},
\end{equation}
Then, a distance-based weighting is applied to these $k$ points as below, \begin{align}
w_{i k} &= \frac{1}{D_{i j_k} + \varepsilon}, 
\quad
\tilde{w}_{i k} = \frac{w_{i k}}{\displaystyle\sum_{m=1}^k w_{i m}},
\end{align}followed by normalization to determine their contributions to the virtual point’s feature, as shown in Fig. \ref{fig:KNN}. 

\begin{equation}
\mathbf{F}_i^{\mathrm{virtual}}
= \sum_{k=1}^K \tilde{w}_{i k}\,\mathbf{F}_{\mathrm{j_k}},
\quad
\
\end{equation}
here, $D_{ij}$ denotes the distance between the virtual point and the original point, and $ {w}_{i k}\ $ refers to the weight of the k original points based on the distance, where $ \tilde{w}_{i k}\ $is the outcome after normalization of the former. The term $\mathbf{F}_i^{\mathrm{virtual}}$ represents the feature of the virtual point, where $\mathbf{F}_{\mathrm{j_k}}$ represents the feature of the original points, and $\varepsilon$ is a small factor to prevent division by zero. 

The final representation of each virtual point integrates its predicted coordinates, features, and logits. 
\begin{equation}
\mathbf{V}_i = \operatorname{Concat}\bigl(\mathbf{v}_i,\;\mathbf{F}_i^{\mathrm{virtual}},\;\mathbf{l}_i^{\mathrm{virtual}}\bigr)
\;\in\;\mathbb{R}^{3 + d + K}\,.
\end{equation}
Finally, we combine the original foreground points, including their coordinates, features, and logits, with the set of virtual points to form a dense point cloud representation.


\subsection{Feature Extraction}\label{feature}

In Section \ref{FPG}, by integrating the point cloud and logits, we obtained a new point cloud map $P^{N \times (C+K)}$ with the logits dimension. In this part, we will further utilize the logits features to improve the feature extraction for more robust representations. 

\begin{figure}[!t]
  \centering
  \includegraphics[width=\columnwidth]{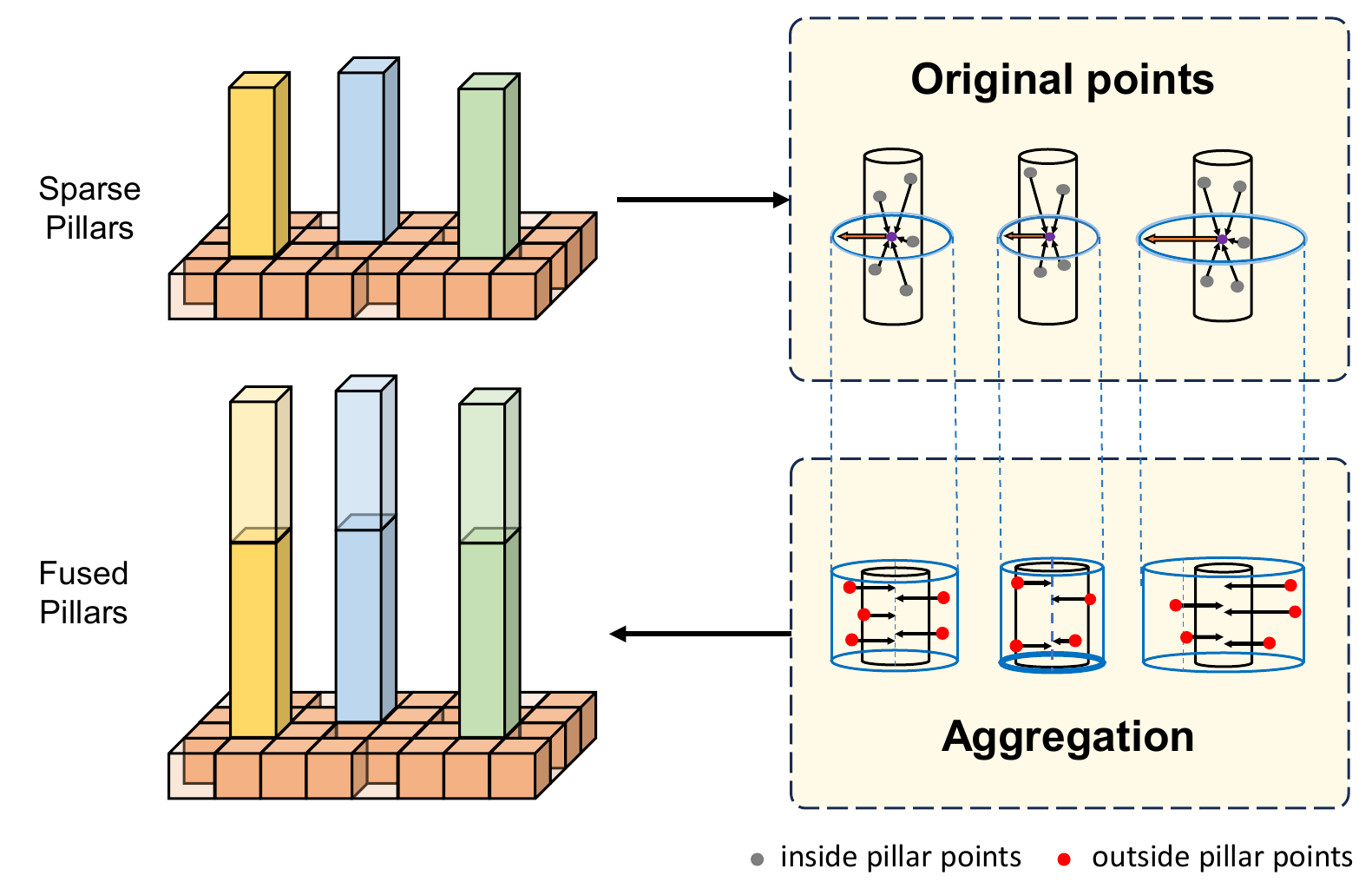}
  \caption{Here are the details about LQE. We firstly computes the aggregation radius (R) based on the inside points, then aggregate the features of the outside points within this radius (R). Ultimately, the updated pillar features are obtained.}
  \label{fig:LQE}
\end{figure}

Specifically, the densified point cloud is taken as input, we perform the logits-aware pillarization. The output is a tensor of size $(D,P,N)$, where \textit{P }is the number of non-empty pillars and \textit{N }is the number of points per pillar. Then, after the pillar encoder \cite{RCFusion,PointPillars}, pillars are transformed to $(C,P)$, where \textit{C }is the number of channels. Then, we devise a logits-query encoder (LQE) , as shown in Fig. \ref{fig:LQE}, innovatively implementing an adaptive radius, enabling each pillar to selectively aggregate the features of its neighboring points based on this radius, enhancing the contextual information of the pillars, thereby improving the robustness. Specifically, we predefine weight values for each category and calculate the number of foreground points after filtering out background points. For each pillar $P_i$, we compute the proportion of points belonging to each category $r_{i_{c}}$ within the pillar and multiply this proportion by the pre-defined weight value $\text{W} $ for the respective category. This determines the contribution of each point within the pillar to the absorption radius based on logits. After determining the absorption radius, we enable each pillar to incorporate the features of its neighboring pillars according to this radius, thus enhancing the contextual information of the pillar features.
\begin{equation}
\text{r}_{i, c} = \frac{\text{N}_{i,c}}{\text{N}_{i,fore}},
\end{equation}

\begin{equation}
\text{W} = [r_{c_{1}}, r_{c_{2}}, r_{c_{3}}],
\end{equation}

\begin{equation}
\text{S}_i = \sum_{c=0}^{2} \text{r}_{i, c} \times \text{W}_{c},
\end{equation}

\begin{equation}
\text{R}_i =    \begin{cases}  \text{S}_i & \text{if } \text{N}_{i,fore} > 0 \\  0.2 & \text{if } \text{N}_{i,fore} = 0   \end{cases},
\end{equation}
Here, $i$ denotes the batch number and $c$ represents the category. $\text{N}_{i,c}$ and $\text{N}_{i,fore} $ respectively denote the quantity of points of category c and the number of foreground points within the i-th column. $r_{c_{i}}$ corresponds to the pre-defined weight radius for each class. $\text{S}_i$ represents the sum of the products of each category contribution and its corresponding weight value, while $\text{R}_i$  indicates the absorption radius of the i-th pillar.

Subsequently, the ball query retrieves the neighboring points around each pillar in accordance with the computed radius. An effective mask is defined to ascertain whether the adjacent pillars have neighbor points. Then, the features and foreground probabilities of the neighbor points are aggregated, and the ultimate neighboring point features are obtained after undergoing max pooling. The neighboring point features and the original pillar features $\mathbf{F}_{\text{pillar}}$ from the pillar encoder are superimposed to acquire the final features $\mathbf{F}'_{\text{pillar}}$, which can be expressed as

\begin{equation}
\mathcal{Q}_i = \{\mathbf{p}_j \mid \|\mathbf{p}_i - \mathbf{p}_j\| \leq R_i, j \in \mathcal{P}\},
\end{equation}

\begin{equation}
\mathbf{F}'_{\text{pillar}}
=
\mathbf{F}_{\text{point}}
+
\mathbf{F}_{\text{pillar}}
+
\mathtt{MLP}\bigl(\bigl[\mathbf{F}_{\text{point}},\,\mathbf{F}_{\text{pillar}}\bigr]\bigr),
\end{equation}
where $\mathcal{Q}_i$ denotes the set of neighboring points surrounding 
pillar $i$, $\|\mathbf{p}_i - \mathbf{p}_j\|$ represents the Euclidean distance between pillar $i$ and point $j$, $\mathbf{F}_{\text{point}}$ indicates the aggregated feature, MLP(\textit{·}) denotes multilayer perceptrons, and the [\textit{·}\textit{, }\textit{·}] denotes the concatenation operation along the channel dimension.
The amplified features are then passed through a sparsely embedded convolutional detection backbone \cite{SECOND}, producing a dense BEV feature map for robust detection. 
\begin{table*}[ht]
    \belowrulesep=0pt
    \aboverulesep=0pt
    \centering
    \footnotesize
    \caption{Comparison with state-of-the-art approaches on the validation set of VoD \cite{VoD}. In the Modality column, R denotes 4D radar and C denotes camera. The best values of the single 4D radar modality are emboldened.}
	\renewcommand\arraystretch{1.2} 
    \begin{tabular}{c|c|cccc|cccc|c}
        \toprule[1.0pt]
        \multirow{2}{*}[-0.7ex]{\centering Method} & \makebox[0.75cm]{\multirow{2}{*}[-0.7ex]{\centering Modality}} & \multicolumn{4}{c|}{Entire Annotated Area AP (\%) $\uparrow$} & \multicolumn{4}{c|}{Driving Corridor AP (\%) $\uparrow$} & \multirow{2}{*}[-0.7ex]{\centering FPS $\uparrow$} \\ 
        \cline{3-6} \cline{7-10}
         &  & \makebox[0.5cm]{Car} & \makebox[0.5cm]{Pedestrian} & \makebox[0.5cm]{Cyclist} & \makebox[0.5cm]{mAP} & \makebox[0.5cm]{Car} & \makebox[0.5cm]{Pedestrian} & \makebox[0.5cm]{Cyclist} & \makebox[0.5cm]{mAP} \\ \midrule
        \text{ImVoxelNet (WACV 2022) \cite{ImVoxelNet}} & C & 19.35 & 5.62 & 17.53 & 14.17 & 49.52 & 9.68 & 28.97 & 29.39 & 7.3 \\
        \midrule 
        \text{FUTR3D (CVPR 2023) \cite{FUTR3D}} & R+C & 46.01 & 35.11 & 65.98 & 49.03 & 78.66 & 43.10 & 86.19 & 69.32 & 11.0 \\
        \text{BEVFusion (ICRA 2023) \cite{BEVFusion}} & R+C & 37.85 & 40.96 & 68.95 & 49.25 & 70.21 & 45.86 & 89.48 & 68.52 & 7.1\\
        \text{RCFusion (IEEE TIM 2023) \cite{RCFusion}} & R+C & 41.70 & 38.95 & 68.31 & 49.65 & 71.87 & 47.50 & 88.33 & 69.23 & 9.0\\
        \text{RCBEVDet (CVPR 2024) \cite{RCBEVDet}} & R+C & 40.63 & 38.86 & 70.48 & 49.99 & 72.48 & 49.89 & 87.01 & 69.80 & -\\
        \text{PointAugmenting (CVPR 2021) \cite{PointAugmenting}} & R+C & 39.62 & 44.48 & 73.70 & 52.60 & 71.02 & 48.59 & 87.57 & 69.06 & 7.9\\
        \text{LXL (IEEE TIV 2023) \cite{LXL}} & R+C & 42.33 & 49.48 & 77.12 & 56.31 & 72.18 & 58.30 & 88.31 & 72.93 & 6.1\\
        \text{SGDet3D (IEEE RAL 2025)} \cite{SGDet3D} & R+C & 53.16 & 49.98 & 76.11 & 59.75 & 81.13 & 60.91 & 90.22 & 77.42 & 9.2 \\
        \midrule 
        \text{PillarNeXt (CVPR 2023) \cite{PillarNeXt}} & R & \text{30.81}  & \text{33.11}& \text{62.78}& \text{42.23}& \text{66.72}& \text{39.03}& \text{85.08}& \text{63.61}& \text{31.6}  \\
        \text{PointPillars (CVPR 2019) \cite{PointPillars}} & R & \text{37.06}& \text{35.04}& \text{63.44}& \text{45.18}& \text{70.15}& \text{47.22}& \text{85.07}& \text{67.48}& \textbf{101.4}  \\
        \text{CenterPoint (CVPR 2021) \cite{centerpoint}} & R & \text{32.74}& \text{38.00}& \text{65.51}& \text{45.42}& \text{62.01}& \text{48.18}& \text{84.98}& \text{65.06}& \text{38.3}  \\
        \text{RadarPillarNet (IEEE T-IM 2023) \cite{RCFusion}}& R & \text{39.30}& \text{35.10}& \text{63.63}& \text{46.01}& \text{71.65}& \text{42.80}& \text{83.14}& \text{65.86}& \text{54.6}  \\
        \text{LXL-R (IEEE TIV 2023) \cite{LXL}} & R & \text{32.75}& \text{39.65}& \text{68.13}& \text{46.84}& \text{70.26}& \text{47.34}& \textbf{87.93}& \text{68.51}& \text{44.7}  \\
        \text{SMURF (IEEE TIV 2023) \cite{SMURF}} & R & \text{\textbf{42.31}}& \text{39.09}& \text{\textbf{71.50}}& \text{50.97}& \text{71.74}& \text{50.54}& \text{86.87}& \text{69.72}& \text{30.3}  \\
        \rowcolor{gray!20}\text{SD4R (ours)} & R & 41.04& \textbf{43.41}& 70.98& \textbf{51.81}& \textbf{71.96}& \textbf{52.32}& \text{86.10}& \textbf{70.13}& \text{22.1}  \\
        \bottomrule[1.0pt]
    \end{tabular}
    \label{tab:VoD_result}
\end{table*}

\subsection{Loss Function}
In addition to the 3D object detection loss $\mathcal{L}_\text{det}$ in \cite{RCFusion}, we also use the point segmentation loss $\mathcal{L}_\text{seg}$ and vote regressing loss $\mathcal{L}_\text{vote}$ in \cite{CluB}, to supervise our model. The total loss is formulated as
\begin{equation}
\mathcal{L}_\text{total} = \mathcal{L}_\text{det} + \lambda(\mathcal{L}_\text{seg}+ \mathcal{L}_\text{vote}),
\end{equation}
where the hyperparameters $\lambda$ balances the loss of foreground virtual points generation process. In this work, we set $\lambda=1.0$.


\section{Experiments}
\subsection{Dataset and Evaluation Metrics}
The View-of-Delft (VoD) \cite{VoD} dataset is a pivotal 4D radar dataset for autonomous driving, providing 3D bounding box annotations. Following its official split, VoD includes 5,139 training and 1,296 validation frames from urban scenarios, focusing on cars, pedestrians, and cyclists.

For comparability, we use the Average Precision (AP) metric with class-specific adaptations. For VoD, we establish a dual-mode protocol: $\text{AP}_\text{EAA}$ (Entire Annotated Area) evaluates all annotated objects without spatial constraints, while $\text{AP}_\text{DC}$ (Driving Corridor) focuses on the collision-critical zone defined as $\mathcal{A}_{DC} = \{(x, y, z) \mid -4\;\text{m} < x < 4\;\text{m}, z < 25\;\text{m}\}$ in camera coordinates. We set IoU thresholds of 0.5 for cars and 0.25 for pedestrians and cyclists to balance detection rigor and practical variability.

\subsection{Network Settings and Training Details}
Our pillar-based processing framework maintains architectural consistency across datasets. For the VoD benchmark, we adhere to official guidelines, setting point cloud boundaries at $X \in [0, 51.2]\,\text{m}$, $Y \in [-25.6, 25.6]\,\text{m}$, and $Z \in [-3, 2]\,\text{m}$, with discretization into $0.16\,\text{m} \times 0.16\,\text{m}$ pillars. For voxel feature extraction, we follow \cite{From_Points_to_Parts} using $0.16\,\text{m} \times 0.16\,\text{m} \times 0.24\,\text{m}$ voxel grids. 

The SD4R framework is trained on 2 NVIDIA GeForce RTX 3090 GPUs using the MMDetection3D ecosystem. We employ the AdamW optimizer with a cyclic learning rate policy (peak rate: 0.003), training for 8-epoch cycles for ablation studies and extending to 24-epoch cycles for final performance validation.
 

\begin{table}[t]
    \belowrulesep=0pt
    \aboverulesep=0pt
    \centering
    \scriptsize
    \footnotesize
    \setlength{\tabcolsep}{2.5pt}
    \caption{Comparison of SD4R and the baseline \cite{RCFusion} across different categories on the validation set of VoD \cite{VoD}.}
    \renewcommand\arraystretch{1.2}
    \begin{tabular}{c|c|ccc}
        \toprule[1.0pt]
        \makebox[1.5cm]{\centering Category} 
      & \makebox[1.5cm]{\centering Method} 
      & \makebox[1.2cm]{\centering 3D} 
      & \makebox[1.2cm]{\centering BEV} 
      & \makebox[1.2cm]{\centering AOS} \\
        \midrule
        \multirow{2}{*}[-0.2ex]{Car} 
      & Baseline & 39.31 & \textbf{49.67} & 31.30 \\
      & SD4R     & \textbf{41.04} & 47.47 & \textbf{35.79} \\
        \midrule
        \multirow{2}{*}[-0.2ex]{Pedestrian} 
      & Baseline & 35.07 & 40.47 & 25.94 \\
      & SD4R     & \textbf{43.41} & \textbf{47.37} & \textbf{33.25} \\
        \midrule
        \multirow{2}{*}[-0.2ex]{Cyclist} 
      & Baseline & 63.63 & 66.36 & 52.37 \\
      & SD4R     & \textbf{70.98} & \textbf{72.13} & \textbf{62.15} \\
        \bottomrule[1.0pt]
    \end{tabular}%
    \label{tab:comparison}
\end{table}

\begin{table}[t]
    \belowrulesep=0pt
    \aboverulesep=0pt
    \footnotesize
    \setlength{\tabcolsep}{1.5pt}
    \caption{Effect of each component in our SD4R in terms of mAP on the VoD dataset \cite{VoD}.}
    \renewcommand\arraystretch{1.2} 
    \centering
    \begin{tabular}{c c c | c c c | c}
        \toprule[1.0pt]
        \makebox[0.9cm]{Baseline} 
      & \makebox[0.9cm]{FPG} 
      & \makebox[0.9cm]{LQE} 
      & \makebox[1.1cm]{Car} 
      & \makebox[1.1cm]{Pedestrian} 
      & \makebox[1.1cm]{Cyclist}
      & \makebox[1.1cm]{mAP} \\
        \midrule
        \checkmark & &                & 39.30 & 35.10 & 63.63 & 46.01 \\
        \checkmark & \checkmark &      & 38.59 & 37.17 & 64.16 & 46.64 \\
        \checkmark & \checkmark & \checkmark & \textbf{40.81} & \textbf{42.02} & \textbf{65.30} & \textbf{49.38} \\
        \bottomrule[1.0pt]
    \end{tabular}%
    \label{tab:ablation_all}
\end{table}

\subsection{3D Object Detection Results}
\begin{figure*}[ht]
  \centering
  
  \includegraphics[width=\textwidth]{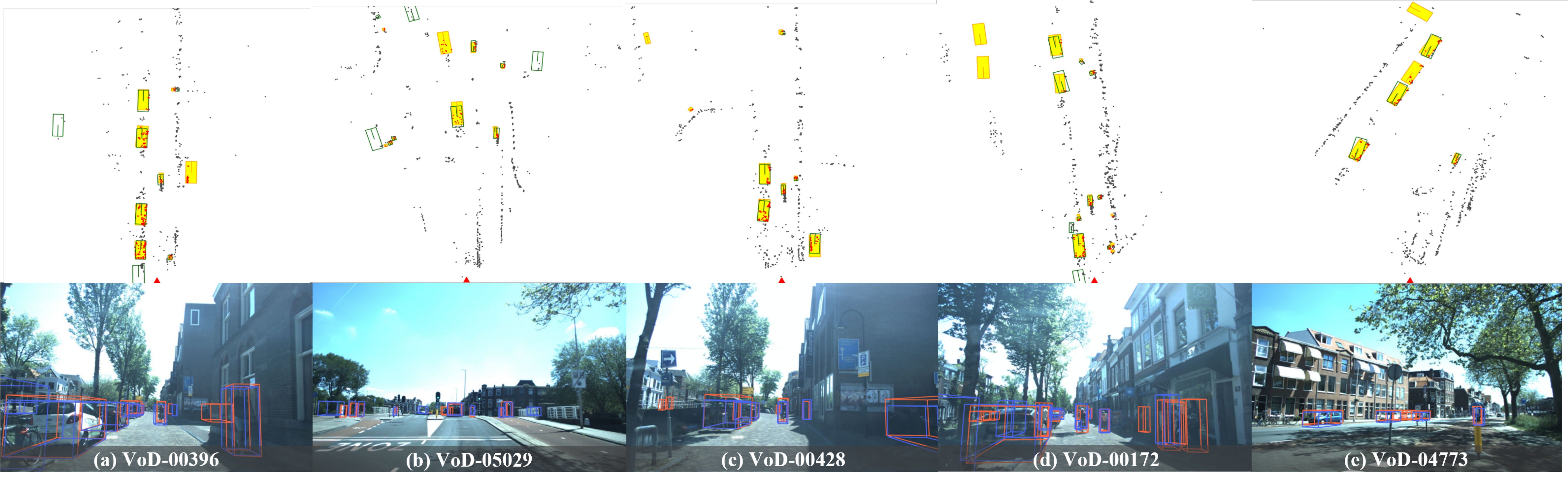}
\caption{Some visualization results on the VoD \cite{VoD} validation. Each column corresponds to a frame of data containing radar points in BEV and an image, where the red triangle denotes the position of the ego-vehicle. Ground-truth boxes are shown in orange (perspective) and yellow (bird’s-eye), while predicted boxes appear in blue and green, respectively. The second row overlays SD4R’s image-plane predictions.}
  \label{fig:outputs}
  \vspace{-2mm} 
\end{figure*}

\textbf{Results on VoD}
As shown in Table \ref{tab:VoD_result}, Our proposed SD4R outperforms all prior approaches in terms of 3D mean Average Precision (mAP) across both the entire annotated area and the driving corridor. This consistently strong performance indicates SD4R’s enhanced resilience to radar-specific challenges such as sparsity and noise. The improvement is particularly evident in the detection of pedestrians and cyclists, which typically suffer from low radar return density and inconsistent signatures. Notably, SD4R also achieves competitive performance on cyclist and car detection, though slightly below the highest mAP. This minor gap can be attributed to the inherent difficulty of capturing motion profiles due to their elongated shapes and the trailing artifacts in radar reflections, factors that can impair spatial feature alignment. Table \ref{tab:VoD_result} also shows the inference speed. SD4R operates at 22.1 frames per second (FPS). While this is more computationally intensive than some other single-modality methods, it remains significantly faster than multi-modal fusion approaches, which typically incur higher overhead from processing multiple data streams. Nonetheless, the runtime remains within acceptable bounds for near real-time applications. Importantly, SD4R closes the performance gap between radar-only and radar-camera fusion models, while also maintaining a high inference speed, offering a competitive, camera-free alternative in adverse weather or lighting conditions where cameras may fail. Visualization of the VoD dataset are presented in Fig. \ref{fig:outputs}.

\textbf{Comparison with Baseline} 
Table \ref{tab:comparison} compares our SD4R framework with the baseline RadarPillarNet \cite{RCFusion} across three metrics: 3D detection accuracy, bird’s-eye view (BEV) precision, and orientation similarity (AOS), highlighting SD4R’s fine-grained advantages. Notably, SD4R achieves significant improvements in pedestrian detection, while gains in car detection are modest. This disparity stems from our logits-guided feature enrichment, which disproportionately enhances performance for categories with sparse and irregular point distributions, such as pedestrians. By weighting and fusing neighboring point features with predicted class probabilities, SD4R amplifies weak signals from small or undersampled objects, improving BEV precision and detection accuracy. Conversely, cars, which form dense and regular pillars under standard encoding, offer limited scope for further improvement. These results demonstrate that leveraging class probabilities for feature aggregation delivers the most substantial benefits in scenarios where baseline representations are weakest.

\begin{table}[t]
    \belowrulesep=0pt
    \renewcommand\arraystretch{1.2} 
    \centering
    \setlength{\tabcolsep}{4pt}
    \footnotesize
    \caption{Ablation of radius parameters within our LQE.}
    \begin{tabular}{ccc|ccc|c}
        \specialrule{0.1em}{0em}{0em}
        \makebox[0.9cm]{$R_p$}
      & \makebox[0.9cm]{$R_{\mathrm{cyc}}$}
      & \makebox[0.9cm]{$R_{\mathrm{car}}$}
      & \makebox[0.9cm]{Ped}
      & \makebox[0.9cm]{Cyc}
      & \makebox[0.9cm]{Car}
      & \makebox[0.9cm]{mAP} \\
        \specialrule{0.05em}{0em}{0em}
        0.2 & 0.2 & 0.2 & 33.94 & 50.67 & 32.92 & 39.18 \\
        \rowcolor{gray!20}
        0.3 & 0.3 & 0.3 & 35.39 & 55.03 & 35.55 & \textbf{41.99} \\
        0.4 & 0.4 & 0.4 & 32.45 & 52.14 & 40.78 & 38.98 \\
        \specialrule{0.05em}{0em}{0em} 
        0.1 & 0.3 & 0.3 & 32.07 & 51.89 & 35.04 & 39.67 \\
        0.2 & 0.3 & 0.3 & 37.06 & 52.59 & 33.16 & 40.94 \\
        0.3 & 0.2 & 0.3 & 34.79 & 54.06 & 34.66 & 41.17 \\
        0.3 & 0.4 & 0.3 & 33.16 & 53.89 & 36.74 & 41.27 \\
        0.3 & 0.3 & 0.4 & 34.54 & 51.44 & 41.98 & 42.65 \\
        0.3 & 0.3 & 0.5 & 32.09 & 51.65 & 40.68 & 41.47 \\
        \rowcolor{gray!20}
        0.2 & 0.3 & 0.4 & 36.84 & 56.79 & 40.46 & \textbf{44.70} \\
        \specialrule{0.1em}{0em}{0em}
    \end{tabular}%
    \label{tab:LQE}
\end{table}

\subsection{Ablation Study}

All ablation studies were conducted on the VoD \cite{VoD} validation set using mean Average Precision (mAP) for three categories (car, pedestrian, cyclist), with results in Table \ref{tab:ablation_all}. Our baseline, RadarPillarNet \cite{RCFusion}, achieves a 46.01\% mAP, with category-specific APs of 39.30\% (car), 35.10\% (pedestrian), and 63.63\% (cyclist).

Integrating the foreground points generator (FPG) module, which generates virtual points to mitigate sparsity, improves the mAP by 0.63\% to 46.64\%. FPG notably boosts pedestrian and cyclist detection by 2.07\% and 0.53\% respectively, while slightly reducing car detection by 0.71\%, indicating its effectiveness for dynamic and sparse targets.

Further incorporating the logits-query encoder (LQE) module, which enhances pillar features by integrating adjacent point features, yields a significant 3.37\% mAP gain over baseline to 49.38\%. LQE demonstrates balanced improvements across all categories, elevating car, pedestrian, and cyclist APs to 40.81\% (+1.51\%), 42.02\% (+4.85\%), and 65.30\% (+1.14\%) respectively. This highlights its role in robust feature extraction.

\textbf{Ablation of LQE} The best performance is achieved when using distinct radius values for each object category, rather than a uniform radius. In particular, assigning a smaller radius for pedestrians and progressively larger radii for cyclists and cars leads to the highest overall accuracy. This aligns with the spatial characteristics of these categories. Pedestrians are small and close-range, requiring finer granularity, while cars occupy larger spatial regions and benefit from broader context. Using tailored radii allows the LQE to better capture relevant neighborhood features for each class, avoiding either oversmoothing or undercoverage. As a result, this category-aware design enhances the effectiveness of semantic enrichment across diverse object categories. 

\section{Conclusions} 
In this work, we address the critical challenges of sparsity and noise in 4D radar point clouds. We propose SD4R, a novel framework that transforms sparse radar point clouds into dense representations, enhancing detection accuracy. By introducing the foreground point generator (FPG), our method effectively mitigates sparsity and noise propagation. Additionally, we design the logit-query encoder (LQE) to enhance pillar representations, resulting in more robust feature extraction. Extensive evaluations on the View-of-Delft dataset \cite{VoD} show that SD4R achieves state-of-the-art performance, highlighting SD4R's potential for transformation from sparse point clouds to dense representations.

\emph{Limitations.} Despite its effectiveness, SD4R has limited inference speed and lacks temporal information. Future work will focus on these issues.

\bibliographystyle{IEEEtran}
\bibliography{reference.bib}

\vspace{12pt}
\color{red}

\end{document}